\documentclass[conference]{IEEEtran}
\IEEEoverridecommandlockouts
\usepackage{cite}
\usepackage{amsmath,amssymb,amsfonts}
\usepackage{algorithmic}
\usepackage{graphicx}
\usepackage{textcomp}
\usepackage{xcolor}
\usepackage{caption}
\usepackage[LGRgreek]{mathastext}

\def\BibTeX{{\rm B\kern-.05em{\sc i\kern-.025em b}\kern-.08em
    T\kern-.1667em\lower.7ex\hbox{E}\kern-.125emX}}
\begin{document}

\title{Leveraging LSTM for Predictive
Modeling of Satellite Clock Bias  
{\footnotesize \textsuperscript{}}

}

\author{\IEEEauthorblockN{ Ahan Bhatt}
\IEEEauthorblockA{\textit{Research Trainee} \\
\textit{Space Applications Centre, ISRO}\\
Ahmedabad, Gujarat \\
bhattahan@gmail.com}
\and
\IEEEauthorblockN{ Ishaan Mehta}
\IEEEauthorblockA{\textit{Research Trainee} \\
\textit{Space Applications Centre, ISRO}\\
Ahmedabad, Gujarat \\
mehtaishaan1234@gmail.com}

\and
\IEEEauthorblockN{Pravin Patidar}
\IEEEauthorblockA{\textit{Sci/Engr "SG"} \\
\textit{Space Applications Centre, ISRO}\\
Ahmedabad, Gujarat \\
pravinpatidar@sac.isro.gov.in}
}

\maketitle

\begin{abstract}
Satellite clock bias prediction plays a crucial role in enhancing the accuracy of satellite navigation systems. In this paper, we propose an approach utilizing Long Short-Term Memory (LSTM) networks to predict satellite clock bias. We gather data from the PRN 8 satellite of the Galileo and preprocess it to obtain a single difference sequence, crucial for normalizing the data. Normalization allows resampling of the data which ensures that the predictions are equidistant and complete. Our methodology involves training the LSTM model on varying lengths of datasets, ranging from 7 days to 31 days. We employ a training set consisting of two days' worth of data in each case. Our LSTM model exhibits exceptional accuracy, with a Root Mean Square Error (RMSE) of $2.11\times 10^{-11}$. Notably, our approach outperforms traditional methods which are used for similar time-series forecasting projects, being 170 times more accurate than RNN, $2.3 \times 10^{7}$ times more accurate than MLP, and $1.9 \times 10^{4}$ times more accurate than ARIMA. This study holds significant potential in enhancing the accuracy and efficiency of low-power receivers used in various devices, particularly those requiring power conservation. By providing more accurate predictions of satellite clock bias, the findings of this research can be integrated into the algorithms of such devices, enabling them to function with heightened precision while conserving power. Improved accuracy in clock bias predictions ensures that low-power receivers can maintain optimal performance levels, thereby enhancing the overall reliability and effectiveness of satellite navigation systems. Consequently, this advancement holds promise for a wide range of applications, including in remote areas, IoT devices, wearable technology, and other devices where power efficiency and navigation accuracy are paramount.

\end{abstract}

\begin{IEEEkeywords}
LSTM, Satellite Navigation, Deep Learning, Clock Bias
\end{IEEEkeywords}

\section{Introduction}
Satellite clock bias prediction is an important aspect of modern Satellite Navigation Systems, influencing the accuracy of location-based services and navigation systems. Accurate prediction of satellite clock bias is crucial for ensuring precise positioning, navigation, and timing in various applications, including aviation, maritime, and terrestrial navigation. Traditional methods for satellite clock bias prediction often face challenges in capturing the complex temporal dependencies inherent in the data.

In recent years, the advent of deep learning techniques, particularly Long Short-Term Memory (LSTM) networks, has offered promising avenues for improving the accuracy and efficiency of satellite clock bias prediction. LSTM networks excel in capturing long-term dependencies in sequential data, making them well-suited for modeling the temporal dynamics of satellite clock bias.

According to multiple studies (Huang et al, 2021[2] and He et al, 2023 [3]) the most suitable method to detect trend terms is through Median Absolute Deviation (MAD). However, we find that since the data is already standardized through a single difference sequence, there is no need to apply Median Absolute Deviation. The data can be easily resampled through the integrated functions in one of the Python libraries i.e. numpy, resulting in faster processing and more accurate results.

In this paper, we present an approach to satellite clock bias prediction utilizing LSTM networks. We aim to leverage the capabilities of LSTM networks to effectively capture the intricate temporal patterns present in satellite clock bias data obtained from the PRN 2 satellite of the Galileo GNSS. Our methodology involves preprocessing the data to obtain a single difference sequence, crucial for normalizing the data to improve prediction accuracy and resampling it for a uniform prediction interval.

Furthermore, we investigate the impact of training data length on prediction performance, considering datasets ranging from 7 days to 31 days in duration. By training the LSTM model on varying lengths of data and evaluating its performance, we seek to determine the optimal testing duration for achieving the highest prediction accuracy. The testing duration is an important factor because LSTM is a recurrent model.

Additionally, we compare the performance of our LSTM-based approach with traditional methods such as Recurrent Neural Networks (RNN), Multi-Layer Perceptrons (MLP), and Autoregressive Integrated Moving Average (ARIMA) models. By quantitatively assessing the accuracy of each method through metrics such as Root Mean Square Error (RMSE), we aim to demonstrate the superiority of LSTM networks in satellite clock bias prediction.

Overall, our research aims to contribute to the advancement of satellite navigation systems by providing a robust and accurate method for predicting satellite clock bias, thereby enhancing the reliability and precision of satellite-based positioning and navigation applications.

\section{Methodology}
The methodology comprises several steps: firstly, data is collected from the PRN 2 satellite of the Galileo system. Subsequently, a single difference sequence is generated from the collected data to facilitate data normalization. Data is resampled according to smallest interval value between two consecutive values in the dataset. The LSTM model is then trained using varying lengths of datasets, ranging from 7 days to 31 days, with a two-day training duration for each dataset. The performance of the LSTM model is evaluated using Root Mean Square Error (RMSE) as the primary metric. Additionally, comparisons are made with similar traditional methods to assess the superiority of the proposed LSTM-based approach in satellite clock bias prediction.

\subsection{Preprocessing}\label{AB}
The preprocessing phase is critical for improving the accuracy of the model's predictions. First, the raw satellite clock bias data is transformed into a single difference sequence to highlight underlying temporal patterns. This transformation removes trends and non-stationary effects, allowing the model to focus on the relevant temporal changes.

Next, resampling is applied to ensure that the data points are equidistant. In this case, we identified that the smallest interval between consecutive data points is 10 minutes. We resample the dataset uniformly to this interval, repositioning existing points and using interpolation techniques where necessary to fill in gaps. By providing equidistant data, we ensure that the LSTM model receives consistent input sequences, which significantly enhances its ability to capture temporal dependencies and make accurate predictions.

 Fig. 1 shows Galileo clock bias data from 1st January-2024 to 31st January-2024.
\begin{figure}
    \centering
    \includegraphics[width=1\linewidth]{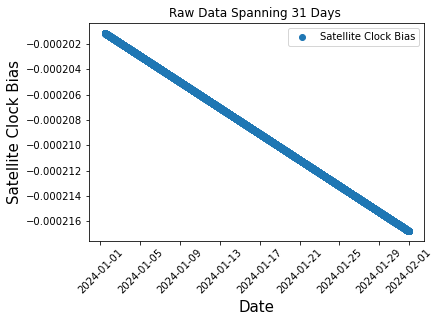}
    \caption{Raw Data}
    \label{fig:enter-label}
\end{figure}
\\ 
\subsubsection{Single Difference Sequence}
The first step in preprocessing the collected data involves transforming it into a single difference sequence. This transformation is essential for accentuating the underlying rare patterns and trends present in the raw satellite clock bias data. By computing the difference between consecutive observations in the dataset, we obtain a sequence that highlights variations and changes over time. This process helps in revealing subtle shifts and fluctuations in the satellite clock bias, which may not be readily apparent in the original data. Moreover, the single difference sequence serves as a precursor to normalization and resampling, as it provides a structured representation of the variating dynamics inherent in the satellite clock bias data.The Fig. 2 shows the Single Different Sequence obtained on the experimental data.
\begin{figure}[h]
    \centering
    \includegraphics[width=1\linewidth]{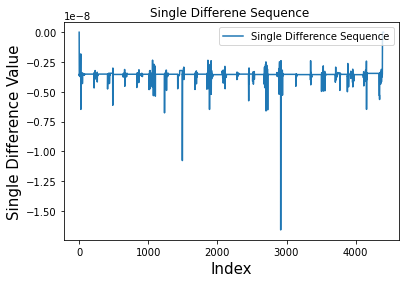}
    \caption{Single Difference Sequence}
    \label{fig:enter-SDS}
\end{figure}
\\
    Here, the variations are clearly visible and the data is more normalized than it was in it's previous iteration. This will allow the model to predict the frequency jumps more accurately 
\\
\subsubsection{Resampling}
Once the single difference sequence is obtained, the next step involves identifying and resampling from the dataset. To ensure the accuracy and consistency of satellite clock bias predictions, it is essential to resample the data so that predictions are equidistant. This resampling process involves adjusting the data points to a uniform interval, which, in our case, is set at 10 minutes – the smallest interval between two consecutive points in the original dataset. We first determine the smallest time interval between any two consecutive data points, which is 10 minutes in this study. The data points are then repositioned to align with the 10-minute intervals. This involves selecting the data points that are closest to these intervals and repositioning them to match exactly. In cases where no data point exists close enough to a 10-minute mark, interpolation techniques are employed to estimate and insert the missing values. This ensures that there are no gaps in the dataset, and each 10-minute interval has a corresponding data point. After resampling, the dataset is reviewed to ensure that all data points are uniformly spaced at 10-minute intervals. This step verifies the consistency and reliability of the resampled data, which is crucial for accurate model training and prediction. By resampling the data to a uniform interval of 10 minutes, we can ensure that our predictive models operate on a consistent and reliable dataset. This uniformity is vital for the effectiveness of Long Short-Term Memory (LSTM) networks and other time-series forecasting methods, as it reduces the complexity and potential errors associated with irregular time intervals. Consequently, this resampling process enhances the overall accuracy and stability of satellite clock bias predictions, thereby contributing to the advancement of GPS navigation systems.

\subsection{LSTM Architecture}
Long Short-Term Memory (LSTM) networks are a type of recurrent neural network (RNN) architecture specifically designed to address the vanishing gradient problem and capture long-term dependencies in sequential data. Unlike traditional RNNs, which suffer from difficulties in retaining and propagating information over long sequences, LSTMs incorporate specialized mechanisms, such as input, forget, and output gates, to regulate the flow of information through the network. This segment provides an overview of the LSTM architecture and its key components, focusing on the functionality of the input gate, forget gate, and output gate. Fig. 3 shows the architecture of a singular LSTM layer.
\begin{figure}[h]
    \centering
    \includegraphics[width=1\linewidth]{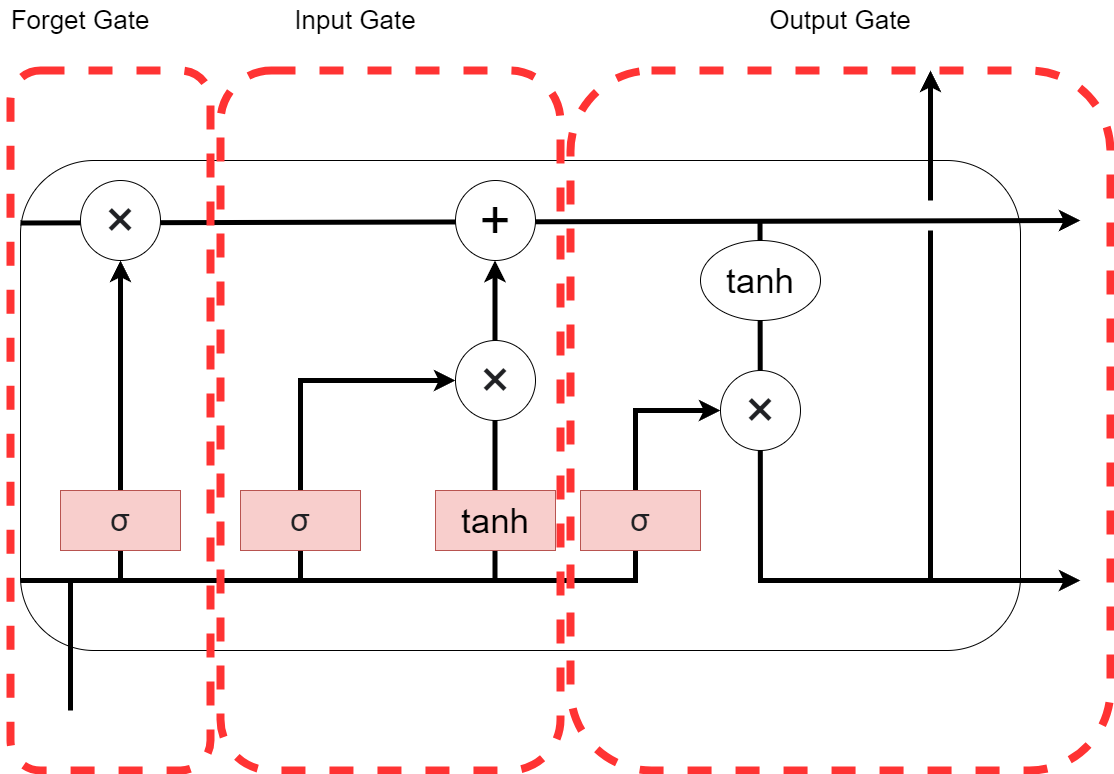}
    \caption{LSTM Model Architecture}
    \label{fig:LSTM Architecture}
\end{figure}\\

\subsubsection{Input Gate}
The input gate in an LSTM network controls the flow of new information into the memory cell. It consists of a sigmoid activation function that takes as input the current input and the previous hidden state. The sigmoid function outputs values between 0 and 1, representing the extent to which new information should be added to the cell state. Additionally, a tanh activation function is applied to the input to regulate its scale. The output of the input gate is then combined with the cell state through element-wise multiplication, allowing the network to selectively update the memory based on the relevance of the new input.\\

\subsubsection{Forget Gate}
The forget gate is responsible for determining which information to discard from the memory cell. Similar to the input gate, it employs a sigmoid activation function to generate values between 0 and 1. These values serve as 'forget factors' that determine the degree to which each element of the cell state should be retained or discarded. By selectively resetting certain components of the cell state, the forget gate enables the network to filter out irrelevant information and focus on retaining long-term dependencies that are essential for accurate prediction.\\

\subsubsection{Output Gate}
The output gate controls the flow of information from the memory cell to the network output. Like the input and forget gates, it utilizes a sigmoid activation function to regulate the flow of information. Additionally, it incorporates a tanh activation function to normalize the cell state. The output gate produces a filtered version of the cell state, which is then passed through the sigmoid function to generate the final output of the LSTM cell. By modulating the output based on the relevance of the information stored in the memory cell, the output gate allows the network to produce accurate predictions while effectively capturing long-term dependencies in the sequential data.

\subsection{Model Training}
For training the predictive model, we adopt a Sequential LSTM architecture tailored for satellite clock bias prediction.We chose Long Short-Term Memory (LSTM) networks for this study because they are well-suited to handling the complex temporal dependencies present in satellite clock bias data. Unlike traditional time-series forecasting models such as ARIMA or basic Recurrent Neural Networks (RNN), LSTMs incorporate a gating mechanism (input, forget, and output gates) that allows them to retain relevant information over long sequences while discarding irrelevant data. This feature is essential for capturing both short-term fluctuations and long-term trends in satellite clock bias, where traditional models often struggle. 
In our experiments, LSTM consistently outperformed ARIMA, MLP, and RNN models in terms of accuracy, as measured by Root Mean Square Error (RMSE). Specifically, The LSTM model achieved an RMSE of 2.11 $\times$ 10$^{-11}$, demonstrating its superior performance. The model comprises two LSTM layers with 512 and 256 units respectively, allowing the network to capture complex temporal dependencies present in the input data. Following the LSTM layers, the model incorporates four Dense layers to perform feature extraction and prediction tasks. Three of these Dense layers utilize the Scaled Exponential Linear Unit (SELU) activation function, known for its ability to facilitate self-normalization and improve convergence. The final Dense layer employs a linear activation function to produce continuous output predictions.

During training, we utilize a training dataset consisting of four days' worth of data to train the model. This choice of training duration strikes a balance between computational efficiency and capturing sufficient temporal information for accurate prediction. The model is trained using the Adam optimizer with a learning rate of 0.001, and the mean squared error loss function is employed to quantify the disparity between predicted and actual satellite clock bias values. To prevent overfitting, we implement early stopping with a patience parameter of 3 epochs, halting training if the validation loss fails to improve for consecutive epochs.

We conduct model training for a total of 10 epochs, iteratively adjusting the network parameters to minimize prediction error and enhance generalization performance. Throughout the training process, we monitor both training and validation performance metrics, such as mean squared error and Root Mean Square Error (RMSE), to assess the model's convergence and prevent overfitting. By optimizing the LSTM architecture and training parameters, we aim to develop a robust and accurate predictive model capable of effectively forecasting satellite clock bias with high precision and reliability.

\section{Results and Discussion}
The results obtained from the experimentation and subsequent analysis provide valuable insights into the effectiveness and performance of the proposed LSTM-based approach for satellite clock bias prediction. In this section, we present a comprehensive overview of the empirical findings, elucidating the predictive capabilities of the trained model across varying lengths of training datasets and comparison with traditional methods. Through rigorous evaluation and comparison of prediction accuracy metrics, including Root Mean Square Error (RMSE), we elucidate the superiority of the LSTM architecture over other conventional techniques. Furthermore, we delve into the impact of prediction length on prediction accuracy, shedding light on the optimal training duration for achieving optimal performance. The discussion aims to provide a nuanced understanding of the factors influencing prediction accuracy and highlights the potential implications of the findings for enhancing the reliability and precision of satellite navigation systems.

The performance of the LSTM-based predictive model was evaluated across varying lengths of training datasets, ranging from 7 days to 31 days. The model's predictive capabilities were assessed through comparison of true values versus predicted values, as depicted in Figure  for 31 days, and Table I shows RMSE, MAE, and MAPE values for each time frame.

\begin{table}[h]
\centering
\caption{RMSE, MAE, MAPE for each time frame}
\label{tab:metrics}

\def\arraystretch{2.5}%
\begin{tabular}{|c|c|c|c|}
\hline
\textbf{Time Frame} & \textbf{RMSE} & \textbf{MAE} & \textbf{MAPE} \\
\hline
31 days & $2.11\times 10^{-11}$ & $1.76\times 10^{-11}$ & $0.49$ \\ 
21 days & $2.31\times 10^{-11}$ & $2.02\times 10^{-11}$ & $0.57$ \\ 
14 days & $1.04\times 10^{-11}$ & $8.38\times 10^{-12}$ & $0.23$ \\ 
7 days & $2.11\times 10^{-10}$ & $2.07\times 10^{-10}$ & $5.82$ \\  
\hline
\end{tabular}
\end{table}

Furthermore, to assess the prediction error, the model's performance for the 31 days time frame is visualized in Fig. 4.

These results showcase the model's ability to accurately predict satellite clock bias across different time frames, with the 31 days duration exhibiting the lowest prediction error as shown in Fig. 5.

The comparison of true and predicted values illustrates the model's capability to capture the underlying patterns and dynamics of satellite clock bias, while the table provides quantitative metrics to gauge the accuracy and performance of the predictions across different time frames. These results underscore the efficacy of the LSTM-based approach in satellite clock bias prediction and its potential for enhancing the precision and reliability of satellite-based navigation systems.

\begin{figure}[h!]
    \centering
    \includegraphics[width=1\linewidth]{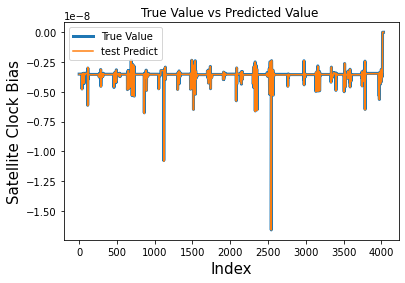}
    \caption{True vs Predicted Values for 31 days}
    \label{fig:2weeks}
\end{figure}

\begin{figure}[h!]
    \centering
    \includegraphics[width=0.5\linewidth]{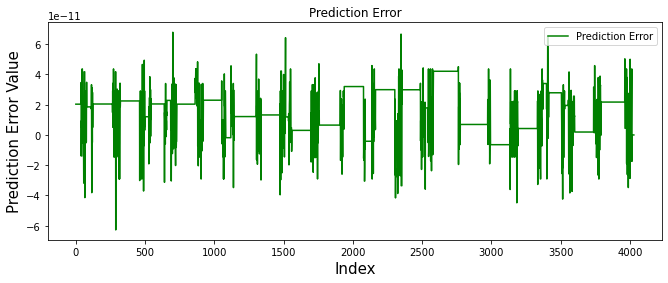}
    \caption{Prediction Error for 31 days}
    \label{fig:error}
\end{figure}
\vfill

\subsubsection{Comparing the LSTM with Other Models}

We compare the performance of the LSTM model with other traditional forecasting models, including ARIMA, MLP, and RNN, across different time frames. Table II, III, IV, V present the Root Mean Squared Error (RMSE), Mean Absolute Error (MAE), and Mean Absolute Percentage Error (MAPE) values for each model and time frame.

\begin{table}[h!]
\centering
\caption{Comparison of Model Performance for 31 days}
\label{tab:comparison_1month}
\def\arraystretch{2.5}%
\begin{tabular}{|c|c|c|c|}
\hline
\textbf{Model} & \textbf{RMSE} & \textbf{MAE} & \textbf{MAPE (\%)} \\
\hline
ARIMA & $4.1\times 10^{-7}$ & $1.81\times 10^{-8}$ & $4.65\times 10^{5}$ \\
LSTM & $2.11\times 10^{-11}$ & $1.76\times 10^{-11}$ & $0.49$ \\
MLP & $0.50$ & $0.50$ & $1.76\times 10^{13}$ \\
RNN & $3.64\times 10^{-9}$ & $3.62\times 10^{-9}$ & $4.62\times 10^{6}$ \\
\hline
\end{tabular}
\end{table}

\begin{table}[h!]
\centering
\caption{Comparison of Model Performance for 21 days}
\label{tab:comparison_3weeks}
\def\arraystretch{2.5}%
\begin{tabular}{|c|c|c|c|}
\hline
\textbf{Model} & \textbf{RMSE} & \textbf{MAE} & \textbf{MAPE (\%)} \\
\hline
ARIMA & $4.72\times 10^{-7}$ & $2.39\times 10^{-8}$ & $1.19\times 10^{6}$ \\
LSTM & $2.31\times 10^{-11}$ & $2.02\times 10^{-11}$ & $0.57$ \\
MLP & $0.41$ & $0.41$ & $7.05\times 10^{13}$ \\
RNN & $1.03\times 10^{-9}$ & $4.37\times 10^{-10}$ & $1.12\times 10^{7}$ \\
\hline
\end{tabular}
\end{table}

\begin{table}[t]
\centering
\caption{Comparison of Model Performance for 14 days}
\label{tab:comparison_2weeks}
\def\arraystretch{2.5}%
\begin{tabular}{|c|c|c|c|}
\hline
\textbf{Model} & \textbf{RMSE} & \textbf{MAE} & \textbf{MAPE (\%)} \\
\hline
ARIMA & $5.63\times 10^{-7}$ & $3.14\times 10^{-8}$ & $1.91\times 10^{6}$ \\
LSTM & $1.04\times 10^{-11}$ & $8.38\times 10^{-12}$ & $0.23$ \\
MLP & $0.65$ & $0.65$ & $3.14\times 10^{14}$ \\
RNN & $9.31\times 10^{-10}$ & $4.26\times 10^{-10}$ & $1.21\times 10^{7}$ \\
\hline
\end{tabular}
\end{table}

\begin{table}[h!]
\centering
\caption{Comparison of Model Performance for 7 days}
\label{tab:comparison_1week}
\def\arraystretch{2.5}%
\begin{tabular}{|c|c|c|c|}
\hline
\textbf{Model} & \textbf{RMSE} & \textbf{MAE} & \textbf{MAPE (\%)} \\
\hline
ARIMA & $4.1\times 10^{-7}$ & $1.81\times 10^{-8}$ & $4.65\times 10^{5}$ \\
LSTM & $2.11\times 10^{-11}$ & $1.76\times 10^{-11}$ & $0.49$ \\
MLP & $0.53$ & $0.04$ & $2.88\times 10^{14}$ \\
RNN & $3.94\times 10^{-9}$ & $3.81\times 10^{-9}$ & $5.69\times 10^{7}$ \\
\hline
\end{tabular}
\end{table}

We must also take a look at the prediction errors of various models. The prediction error is the difference between the actual value and the predicted value of a model. It allows us to clearly view the difference in accuracy between the various models. The prediction errors for various time frames of all the respective models are shown in Fig. 6. To gain a comprehensible understanding, we remove the model with the maximum variations, i.e ARIMA in Fig. 7.

\begin{figure}[h]
\begin{minipage}[h]{0.47\linewidth}
\begin{center}
\includegraphics[width=1\linewidth]{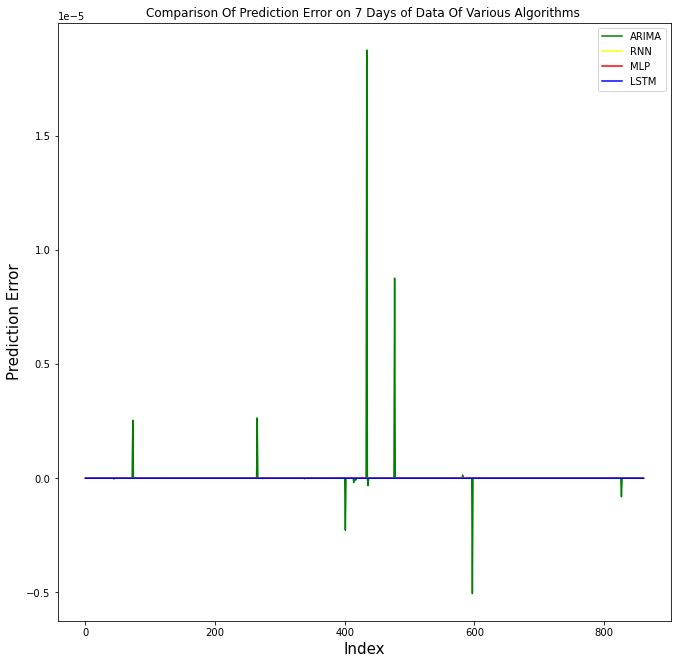} 
\caption*{Comparison of the prediction errors for 7 days}
\end{center} 
\end{minipage}
\hfill
\vspace{0.2 cm}
\begin{minipage}[h]{0.47\linewidth}
\begin{center}
\includegraphics[width=1\linewidth]{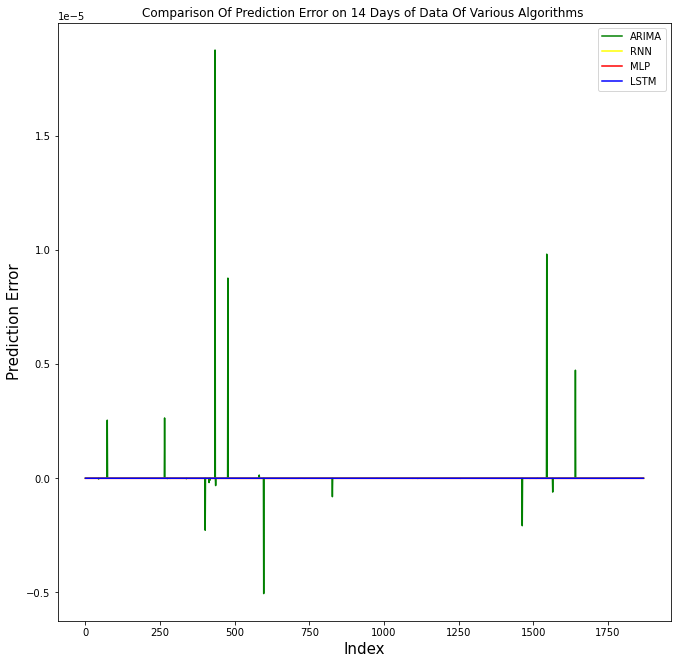} 
\caption*{Comparison of the prediction errors 14 days}
\end{center}
\end{minipage}
\vfill
\vspace{0.2 cm}
\begin{minipage}[h]{0.47\linewidth}
\begin{center}
\includegraphics[width=1\linewidth]{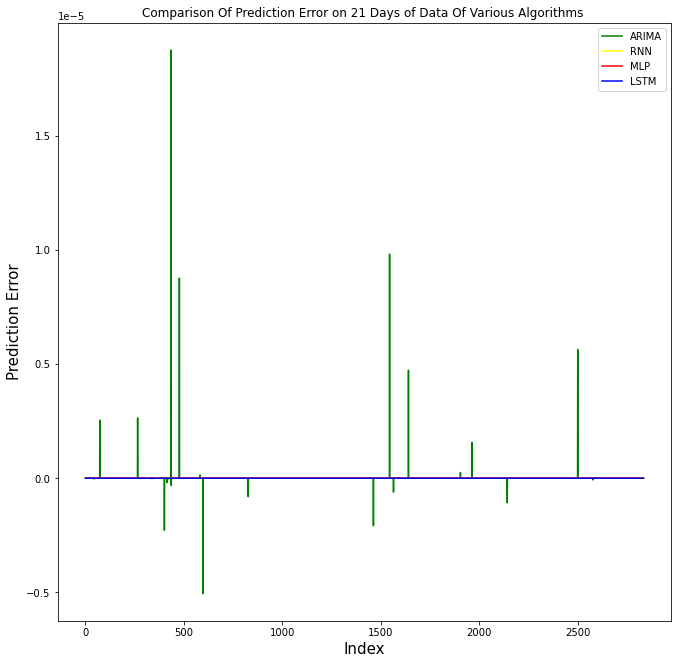} 
\caption*{Comparison of the prediction errors 21 days}
\end{center}
\end{minipage}
\hfill
\begin{minipage}[h]{0.47\linewidth}
\begin{center}
\includegraphics[width=1\linewidth]{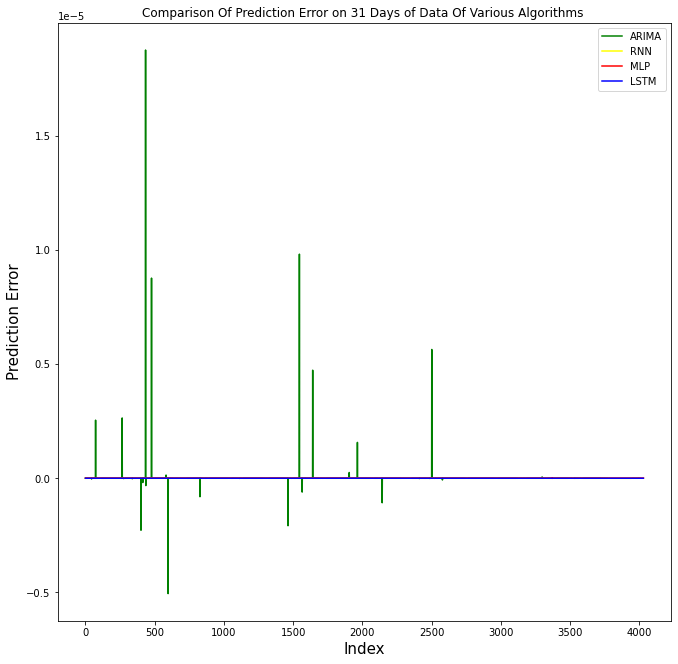} 
\caption*{Comparison of the prediction errors for 31 days}
\end{center}
\end{minipage}
\caption{Comparison of the Prediction Errors of ARIMA, RNN, MLP and LSTM}
\end{figure}

\begin{figure}[h]
\begin{minipage}[h]{0.47\linewidth}
\begin{center}
\includegraphics[width=1\linewidth]{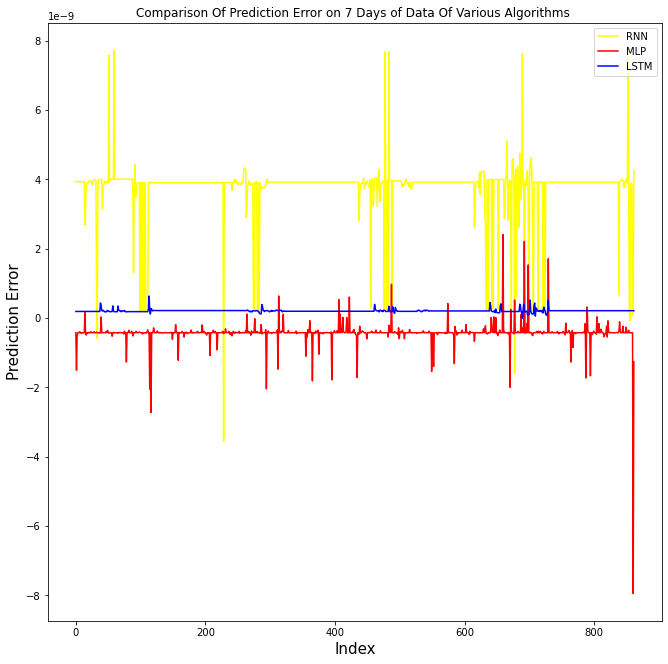} 
\caption*{Comparison of the prediction errors for 7 days}
\end{center} 
\end{minipage}
\hfill
\vspace{0.2 cm}
\begin{minipage}[h]{0.47\linewidth}
\begin{center}
\includegraphics[width=1\linewidth]{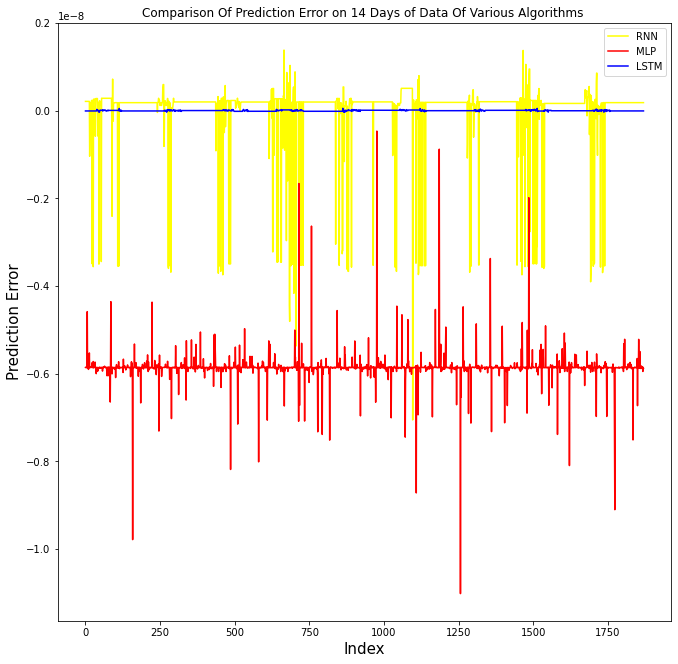} 
\caption*{Comparison of the prediction errors 14 days}
\end{center}
\end{minipage}
\vfill
\vspace{0.2 cm}
\begin{minipage}[h]{0.47\linewidth}
\begin{center}
\includegraphics[width=1\linewidth]{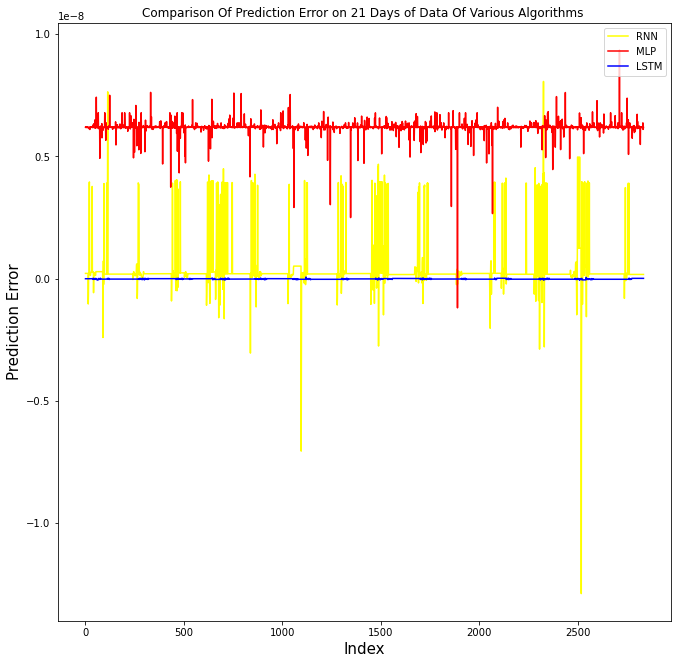} 
\caption*{Comparison of the prediction errors 21 days}
\end{center}
\end{minipage}
\hfill
\begin{minipage}[h]{0.47\linewidth}
\begin{center}
\includegraphics[width=1\linewidth]{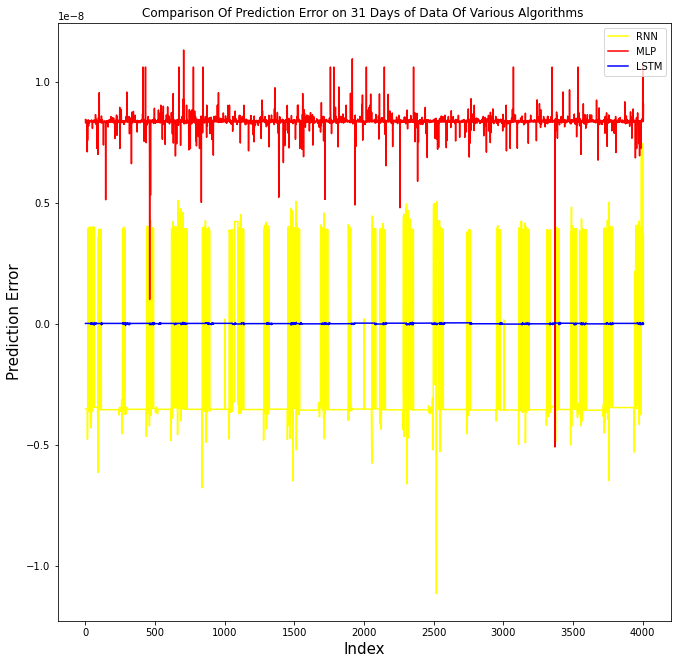} 
\caption*{Comparison of the prediction errors for 31 days}
\end{center}
\end{minipage}
\caption{Comparison of the Prediction Errors of  RNN, MLP and LSTM}
\label{ris}
\end{figure}

\section{Conclusion}
In conclusion, this research paper introduces an approach utilizing Long Short-Term Memory (LSTM) networks for the prediction of satellite clock bias, a critical factor in improving the accuracy of any satellite navigation systems. Traditional methods for satellite clock bias prediction often struggle with capturing the intricate variations such as frequency jumps which are inherent in the data. However, the advent of deep learning techniques, particularly LSTM networks, presents a promising avenue for addressing this challenge.

Through comprehensive experimentation and analysis, we demonstrate the effectiveness of the proposed LSTM-based approach in accurately predicting satellite clock bias. Our methodology involves preprocessing the data, including resampling the data and positioning the closest value to the resampled interval, and training the LSTM model on varying lengths of datasets. The LSTM architecture, with its specialized mechanisms for capturing long-term variations in Time-series data, proves to be highly suitable for modeling the temporal dynamics of satellite clock bias.

Our results indicate that the LSTM model consistently outperforms traditional methods across all time frames. The superior performance of the LSTM model, as evidenced by metrics like Root Mean Square Error (RMSE), Mean Absolute Error (MAE), Mean Absolute Percentage Error (MAPE) underscores its potential for enhancing the precision and reliability of navigation systems.

While the results of this study demonstrate the effectiveness of LSTM in satellite clock bias prediction, it is important to acknowledge the limitations of the dataset. Due to the small size of the dataset, we found that more advanced models, such as TimeGPT and statsforecast, were unable to outperform simpler methods like ARIMA. As a result, these models were excluded from the final analysis.

Furthermore, the confidentiality and unavailability of the dataset have prevented further experimentation and comparison with other models. Despite these limitations, the LSTM model proved highly accurate in this context, providing a strong foundation for future work. Larger datasets or different conditions may allow for more comprehensive comparisons, but the findings here emphasize LSTM’s robustness when data is limited.

The insights gained from this study offer valuable implications for the optimization of low-power receivers utilized in diverse devices requiring power conservation. By leveraging the accurate predictions of satellite clock bias provided by our LSTM-based approach, these devices can operate with enhanced precision without compromising on power efficiency. This advancement not only facilitates the seamless functioning of low-power receivers but also extends the benefits to a myriad of applications, spanning from IoT devices to wearable technology, where both navigation accuracy and power conservation are critical. As such, our research contributes to the advancement of satellite navigation systems, enabling them to deliver reliable and efficient performance across various domains, thereby enhancing user experience and operational efficacy.

\newpage
\section*{Acknowledgment}

This research was conducted at the Space Applications Centre, ISRO, and would not have been possible without the support and resources provided by the organization. We extend our sincere gratitude to the entire team at Space Applications Centre for their encouragement and assistance throughout this project.

We would like to thank the team behind the GPS and Galileo systems for their groundbreaking work and the invaluable data they have made available. Their contributions have been instrumental in advancing satellite navigation technology.

Special thanks to our colleague, Harsh N Bhatt, for his meticulous efforts in refining the data that formed the basis of our research. His expertise and dedication significantly enhanced the quality of our work.

We also express our deep appreciation to Sri. Pravin Patidar, our supervisor and co-author, for his invaluable guidance and support. His leadership and insights were crucial in steering this project to success.

Thank you to all who contributed directly or indirectly to this research. Your efforts and support are greatly appreciated.

\vspace{12pt}

\end{document}